\documentclass[10pt,twocolumn,letterpaper]{article}

\usepackage{iccv}              % To produce the CAMERA-READY version
\usepackage{booktabs}
\usepackage{multirow}
\usepackage{colortbl}
\usepackage[normalem]{ulem}
\usepackage{bbding}
\usepackage[accsupp]{axessibility}  % Improves PDF readability for those with disabilities.
\useunder{\uline}{\ul}{}% To produce the REVIEW version
\definecolor{iccvblue}{rgb}{0.21,0.49,0.74}
\usepackage[pagebackref,breaklinks,colorlinks,allcolors=iccvblue]{hyperref}

\def\paperID{7970} % *** Enter the Paper ID here
\def\confName{ICCV}
\def\confYear{2025}

\title{Synthesizing Near-Boundary OOD Samples for Out-of-Distribution Detection}

\author{Jinglun Li$^{1,3}$, Kaixun Jiang$^1$, Zhaoyu Chen$^1$, Bo Lin$^3$, Yao Tang$^3$, Weifeng Ge$^2$\footnotemark[2], Wenqiang Zhang$^{1,2}$\footnotemark[2]\\
$^1$College of Intelligent Robotics and Advanced Manufacturing, Fudan University, Shanghai\\
$^2$Shanghai Key Lab of Intelligent Information Processing,\\
College of Computer Science and Artificial Intelligence, Fudan University, Shanghai\\
$^3$JIIOV Technology, Beijing\\
{\tt\small \{jinglunli21, kxjiang22\}@m.fudan.edu.cn, zhaoyuchen20@fudan.edu.cn,}\\
{\tt\small \{bo.lin, yao.tang\}@jiiov.com, }\\
{\tt\small weifeng.ge.ic@gmail.com, wqzhang@fudan.edu.cn}
}

\begin{document}
\maketitle
\renewcommand{\thefootnote}{\fnsymbol{footnote}} %将脚注符号设置为fnsymbol类型，即特殊符号表示
\footnotetext[2]{ indicates corresponding authors.}

\begin{abstract}
    Pre-trained vision-language models have exhibited remarkable abilities in detecting out-of-distribution (OOD) samples. However, some challenging OOD samples, which lie close to in-distribution (InD) data in image feature space, can still lead to misclassification. 
    The emergence of foundation models like diffusion models and multimodal large language models (MLLMs) offers a potential solution to this issue. In this work, we propose \textbf{SynOOD}, a novel approach that harnesses foundation models to generate synthetic, challenging OOD data for fine-tuning CLIP models, thereby enhancing boundary-level discrimination between InD and OOD samples.
    Our method uses an iterative in-painting process guided by contextual prompts from MLLMs to produce nuanced, boundary-aligned OOD samples. These samples are refined through noise adjustments based on gradients from OOD scores like the energy score, effectively sampling from the InD/OOD boundary.
    With these carefully synthesized images, we fine-tune the CLIP image encoder and negative label features derived from the text encoder to strengthen connections between near-boundary OOD samples and a set of negative labels. 
    Finally, SynOOD achieves state-of-the-art performance on the large-scale ImageNet benchmark, with minimal increases in parameters and runtime. Our approach significantly surpasses existing methods, and the code is available at https://github.com/Jarvisgivemeasuit/SynOOD.
\end{abstract}

\vspace{-1.5em}
\section{Introduction}
\begin{figure}[!h]
    \centering
    \includegraphics[width=0.9\linewidth]{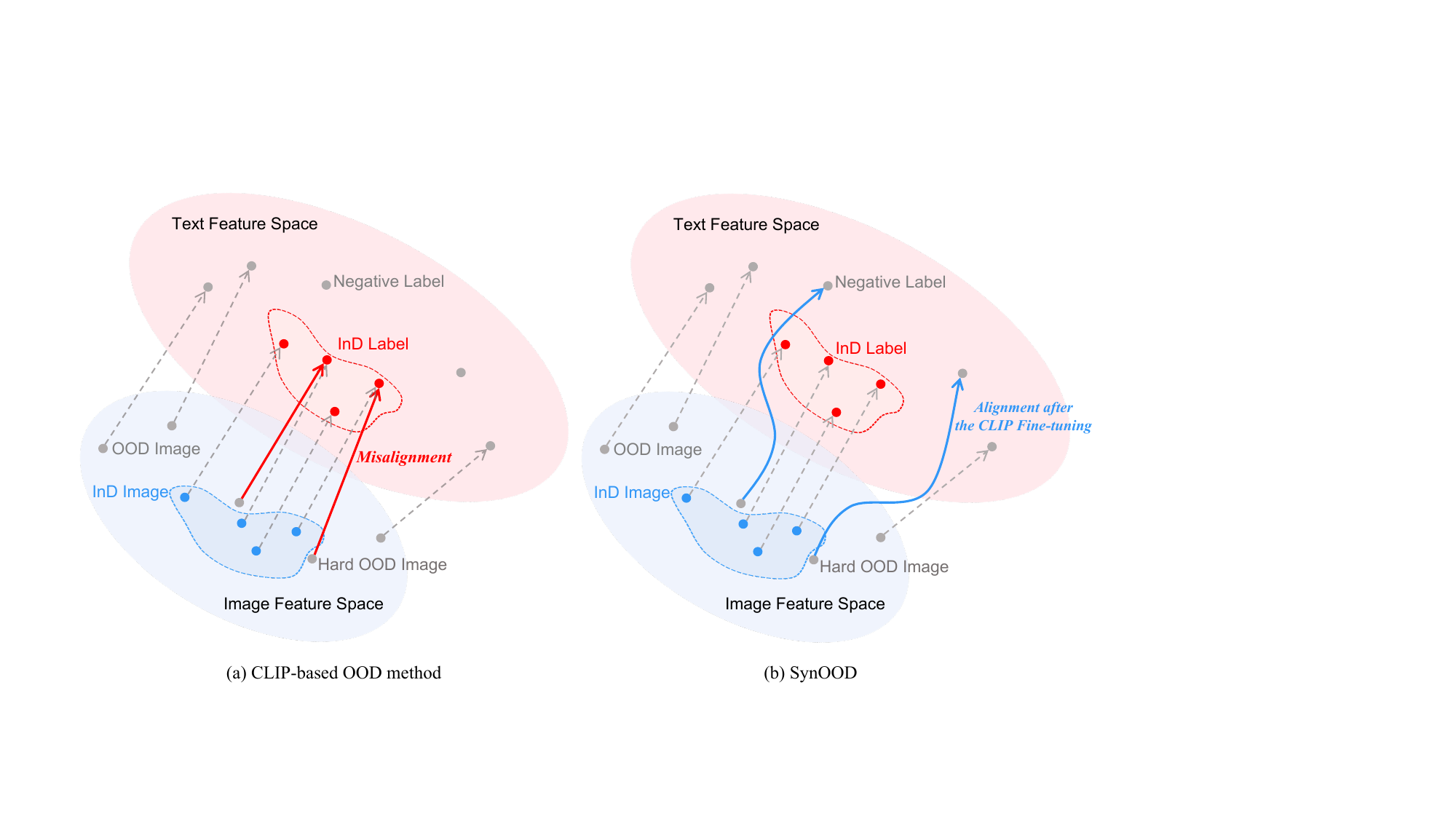}
    \caption{(a) illustrates a simplified example highlighting the limitations of CLIP-based OOD methods, where challenging OOD samples are misclassified due to CLIP models’ limited fine-grained discrimination. (b) Our proposed method addresses this limitation by generating challenging data to fine-tune the CLIP models, building strong connections between confusing OOD samples and their corresponding negative labels.}
    \label{fig:motivation}
    % \vspace{-0.5em}
\end{figure}

 Modern deep neural networks~\cite{liu2021swin, dosovitskiy2021imageworth16x16words, huang2017densely, xie2017aggregated} deployed in open-world scenarios inevitably encounter out-of-distribution (OOD) samples, which can pose significant security risks. Accurate identification of OOD data is crucial to mitigate these threats. Traditional vision-based OOD detection methods~\cite{hendrycks2016baseline, liu2020energy, liang2017enhancing, huang2021importance, Li2023hvcm, sun2022dice, sun2021react} often rely solely on a single image domain. Recent research~\cite{jiang2024negative, li2024tagood, ming2022delving, miyai2024locoop, wang2023clipn} in pre-trained visual-language models~\cite{radford2021learning, li2023blip} has demonstrated significant improvements in OOD detection by effectively employing both visual and language information. In particular some CLIP-based methods~\cite{wang2023clipn, jiang2024negative, ming2022delving}, such as NegLabel~\cite{jiang2024negative}, enhance OOD detection by introducing potential OOD text labels, denoted as negative labels, that lie outside the in-distribution (InD) label space. However, a significant challenge remains in accurately identifying hard OOD samples near the InD/OOD boundary, as these samples often appear visually similar to InD instances, making them difficult to classify using CLIP-based methods directly. CLIP-based methods show that OOD samples situated near the InD/OOD boundary tend to align more closely with InD labels, as images are typically more densely packed in the feature space than labels, limiting the model’s ability to establish clear semantic alignment, this is illustrated in Fig.~\ref{fig:motivation}(a). Consequently, this mismatch reduces the reliability of CLIP~\cite{radford2021learning} in detecting boundary OOD samples, particularly those closely resembling the InD distribution.

A promising approach to improve OOD detection is to effectively map ambiguous samples near the InD/OOD boundary to either InD or negative labels. However, fine-tuning CLIP models for this purpose has been challenging due to a lack of suitable data. Recent advancements in multimodal large language models (MLLMs)~\cite{liu2023llava, qwen, li2023blip, openai2023gpt4} and diffusion models~\cite{rombach2022high, podell2024sdxl} offer powerful generative capabilities, though their application to task-oriented data generation remains relatively unexplored. To fill this gap, we propose a novel iterative generative approach that utilizes the contextual understanding of an MLLM and the sophisticated image synthesis capabilities of a diffusion model. This integration enables the creation of realistic, boundary-aligned OOD samples that are visually similar to InD data while remaining sufficiently distinct. By generating these nuanced near-boundary OOD samples, our approach provides the CLIP model with more challenging data for fine-tuning, achieving a more accurate separation between InD and OOD samples.

Our method begins by using a language model to extract all detected contextual elements within an InD image. For example, in an image labeled “panda,” the language model may detect contextual elements like “bamboo,” “tourist,” “leaf,” and “railing,” which are commonly associated with the primary subject but not central to it. These elements then serve as prompts for an in-painting diffusion model. Rather than relying on predefined masks, we employ an iterative generative process to guide the model in creating images that remain visually similar to the InD data while representing OOD content.
In each iteration, the generated image is evaluated using an OOD detection model, and the resulting OOD score informs a gradient. This gradient is backpropagated through the diffusion model, updating the noise to iteratively adjust the image. Over time, the model gradually replaces primary subject elements, such as the “panda,” with background elements from the identified list. This controlled transformation subtly shifts the image’s focus away from the core theme, allowing it to appear distinct from InD samples without losing its underlying visual similarities. By iteratively integrating contextual elements as the main focus while maintaining the original style and setting, the resulting synthetic images closely resemble InD examples in appearance but remain distinctly OOD, aligning with the theoretical principles in \cite{fang2022out, zheng2023out}.

In this work, we propose \textbf{SynOOD}, a novel method that iteratively generates near-boundary data to fine-tune the CLIP models for enhancing OOD detection performance. Specifically, our method contains three components: an iterative generative process, fine-tuning the CLIP image encoder with a projection layer, and refining negative label features derived from the CLIP text encoder. This process significantly boosts the model’s capacity to distinguish between InD and OOD samples, this is illustrated in Fig.~\ref{fig:motivation}(b). By integrating these processes, SynOOD offers a robust and effective approach to OOD detection. Our contributions are summarized as follows:
\begin{itemize}
    \item We propose \textbf{SynOOD}, a novel framework for OOD detection that generates challenging, near-boundary OOD samples to the fine-tune CLIP models, enhancing to detect difficult OOD cases close to the InD/OOD boundary.
    \item We introduce a generation process that iterative synthesizes near-boundary OOD samples using foundation models, guided by OOD gradient information. This process yields high-quality, nuanced data that enhances CLIP to strengthen connections between challenging OOD samples and negative labels.
    \item Extensive experiments show that SynOOD achieves state-of-the-art performance on widely used large-scale benchmarks, with minimal increases in parameters and runtime, outperforming existing methods by improving AUROC by 2.80\% and reducing FPR95 by 11.13\%.
\end{itemize}

\section{Related Work}
\textbf{OOD detection with visual modal.} Single-modal visual OOD detection methods include: (1) Logit-based approaches, which compute OOD scores from network logits. MSP~\cite{hendrycks2016baseline} uses the maximum logit, while ODIN~\cite{liang2017enhancing} enhances separation via input perturbations and logit rescaling. ReAct~\cite{sun2021react} further reduces overconfidence by adjusting activation logits. (2) Distance-based methods, which use feature distances between InD and OOD samples as OOD scores. Gaussian discriminant analysis~\cite{lee2018simple, winkens2020contrastive} and metrics like cosine similarity~\cite{chen2020boundary, zaeemzadeh2021out}, Euclidean distance~\cite{huang2020feature}, and RBF kernels~\cite{van2020uncertainty} are commonly employed. (3) Gradient-based methods, such as GradNorm~\cite{huang2021importance}, leverage classifier gradients to distinguish InD from OOD samples using gradient-based features.

\noindent\textbf{OOD detection with multi-modal models.}
Leveraging textual information alongside visual data for OOD detection has become increasingly popular due to its strong performance. Fort et al.\cite{fort2021exploring} pioneered this direction by using class names of potential outliers as input to image-text encoders like CLIP, improving OOD detection. MCM\cite{ming2022delving} is an effective post-hoc method that uses the maximum predicted softmax value from a vision-language model as the OOD score. More recently, CLIPN~\cite{wang2023clipn} proposed using a text encoder to identify OOD samples by comparing similarity discrepancies between two text encoders and a frozen image encoder. Building on this, LSN~\cite{nie2024out} introduced negative classifiers with learned prompts to detect images outside specific categories. NPOS~\cite{tao2023non} generates synthetic OOD data to better define decision boundaries between InD and OOD samples. LAPT~\cite{zhang2024lapt} automates prompt tuning for vision-language models, reducing manual effort. DreamOOD~\cite{du2023dream} learns a text-conditioned latent space to generate diverse OOD samples by decoding low-likelihood embeddings into images. NegLabel~\cite{jiang2024negative} selects potential OOD labels from semantically related WordNet~\cite{miller1995wordnet} terms outside the InD label space, using a pre-trained vision-language model like CLIP to classify images as InD or OOD.

\section{Method}
\subsection{OOD Detection Setup}
Given a training set $\mathcal{D}^\text{in}=\{(x_i, y_i)\}^n_{i=1}$, where $x_i\in \mathbb{R}^{3\times H\times W}$ is the 3-channel image of size $H\times W$, $y_i\in\mathcal{Y}$ denotes one of $C$ InD categories, and $n$ is the number of samples, our target is to develop an OOD detector $G(x)$ solely based on $\mathcal{D}^\text{in}$. When applied to a test image set $\mathcal{X}=\{x_i\}^K_{i=1}$, the detector $G(x)$ should output a binary classification using a score function $S(x)$: 
\begin{equation}
    G(x)=\left\{
    \begin{aligned}
        &{\rm InD},  &{\rm if}\quad S(x) \geq \eta; \\
        &{\rm OOD},  &{\rm if}\quad S(x) < \eta,
    \end{aligned}
    \right.
    \label{eq:ood_detector}
\end{equation}
where $\eta$ is a threshold parameter. We follow Jiang et al.~\cite{jiang2024negative} and employ a negative label set $\mathcal{Y}^-=\{y_{C+1},...,y_{C+M}\}$ for classification:
\begin{equation}
    S(x)=\frac{\text{sim}(x, \mathcal{Y})}{\text{sim}(x, \mathcal{Y})+\text{sim}(x, \mathcal{Y}^-))}
\end{equation}
where $\text{sim}(x, \cdot)$ represents the sum of CLIP similarities between the sample and the labels in a given label set. 

\subsection{Overview of SynOOD}
\begin{figure*}[t]
    \centering
    \includegraphics[width=0.9\linewidth]{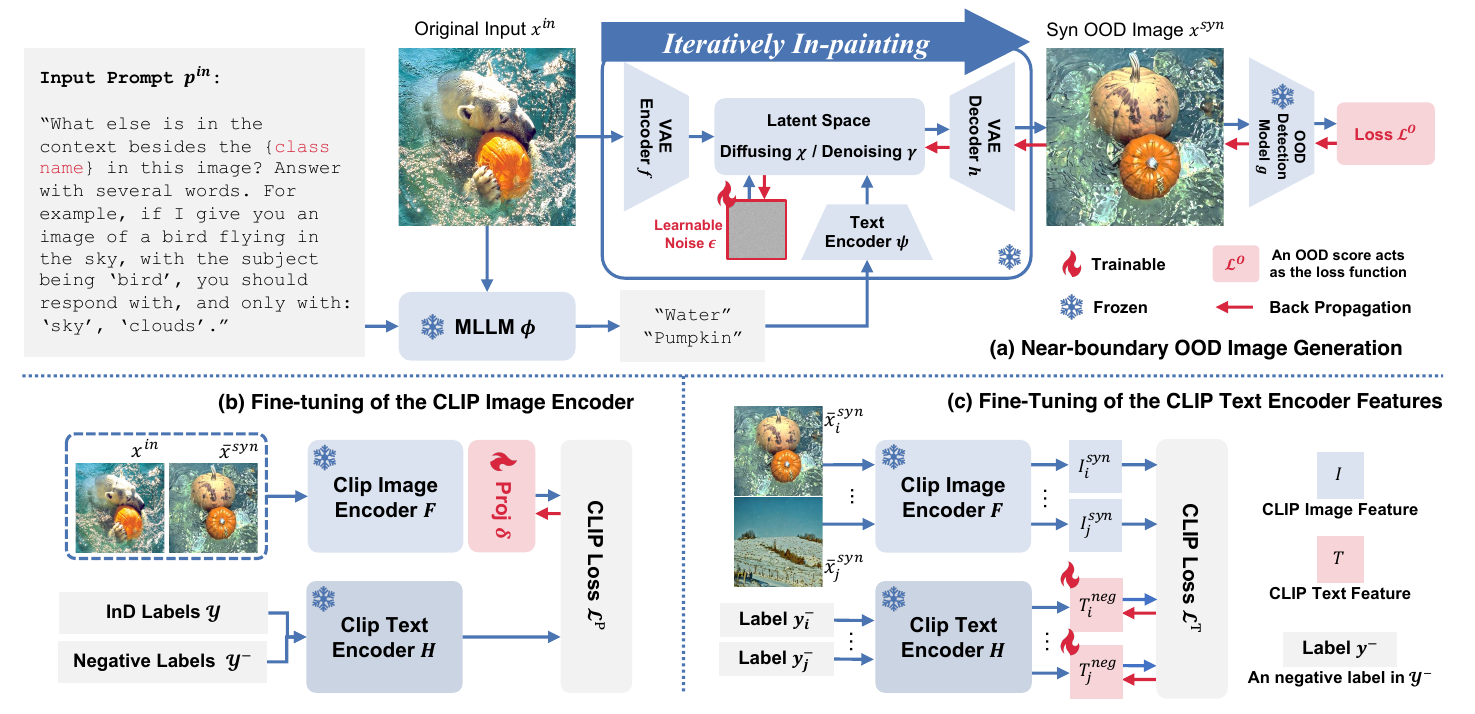}
    \caption{Overview of our proposed SynOOD framework. (a) Near-boundary OOD image generation: utilizes an MLLM and a diffusion model to iteratively generate synthetic OOD images from InD images, guided by an OOD score as the loss function. (b) Fine-tuning of the CLIP image encoder: trains a projection to strengthen connections between challenging OOD samples and negative labels. (c) Fine-tuning of the CLIP text encoder features: refines the negative label features derived from CLIP to improve OOD discrimination further.}
    \label{fig:framework}
    \vspace{-1em}
\end{figure*}

 Our proposed SynOOD, illustrated in Fig.~\ref{fig:framework}, addresses this issue through a three-step process: \textbf{1) Near-Boundary OOD Image Generation.} In Fig.~\ref{fig:framework}(a), an MLLM is employed to generate multiple semantic labels for each element within an image, excluding the main object. A novel iterative generative approach utilizing a diffusion model is then applied to generate near-boundary OOD images. These synthetic images help us to fine-tune the CLIP models effectively. \textbf{2) Fine-tuning of the CLIP image encoder.} We train a projection layer following the CLIP image encoder using both InD data and synthetic OOD images along with the negative labels. The image encoder remains frozen, while only the projection layer updates. \textbf{3) Fine-tuning of the CLIP text encoder features.} We make the features (the output of the CLIP text encoder) associated with a subset of negative labels related to synthetic OOD images learnable and fine-tune them with synthetic images. This step reduces the semantic gap between InD and negative labels to an appropriate distance, improving image-text alignment. 
 
 We fine-tune the CLIP image encoder and text encoder features separately to maintain training stability. Experiments in Table~\ref{tab:strategy} validate the effectiveness of this approach. After the fine-tuning of the CLIP encoders, our approach boosts the performance of OOD detection.
 
\subsection{Near-boundary OOD Image Generation}
\label{subsec:generate}
In this image-generation process, we need to use three models: an MLLM, an in-painting diffusion model, and a traditional recognition model as the OOD detection model. Initially, we employ the MLLM $\phi$ to generate multiple semantic labels for each element in every InD image $x^{\text{in}}$, excluding the main object. The output, denoted as  $p^{\text{con}}$ is obtained as follows:
\begin{equation}
    p^{\text{con}} = \phi(x^{\text{in}}, p^{\text{in}}),
\end{equation}
where $p^{\text{in}}$ is the input prompt for $\phi$. Rather than relying on masks, we implement an iterative generative process when employing an in-painting diffusion model set to a strength of less than 1, enabling the generation of OOD images with minimal manual intervention.

Concretely, $x^{\text{in}}$ and $p^{\text{con}}$ are fed in the in-painting diffusion model to generate an image $x^{\text{syn}}$. Specifically, we denote the feature of $x^{\text{in}}$ as $z^{\text{in}}$ after passing it through the VAE~\cite{kingma2013auto} encoder $f$. Given a learnable random noise $\epsilon$, a variance schedule $\{\alpha_1, ...,\alpha_T\}$, a timestep $T$, and $z^{\text{in}}$, the diffusing process $\chi$ can be expressed as:
% \begin{equation}
%     \label{eq:addnoise}
%     \scalebox{0.9}{$
%     z_T =\chi(z^{\text{in}}, T, \epsilon)= \sqrt{\bar{\alpha}_T}z^{\text{in}}+\sqrt{1-\bar{\alpha}_T}\epsilon, \epsilon \sim \mathcal{N}(0, I).$}
% \end{equation}
\begin{align}
    \label{eq:addnoise}
    z_T &=\chi(z^{\text{in}}, T, \epsilon) \notag\\
    &= \sqrt{\bar{\alpha}_T}z^{\text{in}}+\sqrt{1-\bar{\alpha}_T}\epsilon, \epsilon \sim \mathcal{N}(0, I).
\end{align}

where $\bar{\alpha}_t=\prod_{s=1}^t \alpha_s$. In the denoising process, a U-Net~\cite{ronneberger2015u}, denote as $\epsilon_\theta$, is utilized to predict a noise needed to reconstruct $z_{t-1}$ from $z_t$ using a text prompt $p^{\text{con}}$ and a mask $M$:
\begin{align}
    z_{t-1} =& {\sqrt{\alpha_{t-1}}}(\frac{z_t-\sqrt{1-\alpha_t}\epsilon_\theta(z_t, t, P, M)}{\sqrt\alpha_t})\notag \\
    &+\sqrt{1-\alpha_{t-1}}\epsilon_\theta(z_t, t, P, M),
\end{align}
where $P=\psi(p^{\text{con}})$ stands for the feature of $p^{\text{con}}$ extracted by a text encoder $\psi$, and $M$ is initialized as a matrix filled with ones. 
% Even when the mask $M$ is uniformly set to one, the model utilizes prior information to generate content that closely aligns with the original image’s style and context, making it more suitable than a standard text-to-image model. This approach enables the inpainting model to produce highly similar InD-like images without masking, which is essential for generating near-boundary OOD samples.
By iterating through multiple time steps, the image features are gradually denoised and completed until the image at $t=0$ is completely generated. The synthetic image is obtained through the complete denoising process, represented by $\gamma$:
\begin{align}
\label{eq:denoise}
    x^{\text{syn}}=h(\gamma(z_T, T, P, M)).
\end{align}
where $h$ is the VAE decoder.

We employ an off-the-shelf OOD detection method such as Energy score~\cite{liu2020energy} as a loss function on a traditional recognition model $g$ (e.g. ResNet50~\cite{he2016deep}). The loss function is defined as:
\begin{equation}
\label{eq:loss}
    \mathcal{L}^{O} = m_{\text{out}}-\tau\cdot\text{log}\sum^C_{i=1}e^{g_i(x^{\text{syn}})/\tau},
\end{equation}
where $m_{\text{out}}$ is a constant representing the OOD threshold of the OOD score, as used in~\cite{liu2020energy}, $\tau$ is a temperature parameter, and $g_i(x)$ denotes the logits of $g$ for the $i$-th class among $C$ categories. Combining the Eqs. (\ref{eq:addnoise}), (\ref{eq:denoise}), and (\ref{eq:loss}), we can calculate the gradient of the random noise $\epsilon$ at the very beginning:
\begin{align}
    \nabla_\epsilon \mathcal{L}^{\text{O}}=\frac{\partial L}{\partial x^{\text{syn}}}\cdot\frac{\partial g}{\partial\gamma}\cdot\frac{\partial\gamma}{\partial z_T}\cdot\frac{\partial z_T}{\partial \epsilon}.
\end{align}
By iterative refining $\epsilon$ for a few iterations, we observe rapid convergence of the loss function, leading to the generation of highly reliable near-boundary OOD images $\Bar{x}^{\text{syn}}$.

While calculating the gradient of $\epsilon$ can be computationally demanding, we address this challenge by adopting the Skip Gradient operation proposed by Chen et al.~\cite{NEURIPS2023_skipgrad}:
\begin{align}
    \nabla_\epsilon \mathcal{L}^{\text{O}}\approx\ddot{\nabla}_\epsilon \mathcal{L}^{\text{O}}=\rho\cdot\frac{\partial L}{\partial x^{\text{syn}}}\cdot\frac{\partial g}{\partial\gamma}
\end{align}
This technique significantly reduces the computational burden, enabling more efficient training. The noise updating equation can be expressed as:
\begin{align}
    \epsilon:= \epsilon-r\cdot\ddot{\nabla}_\epsilon \mathcal{L}^{\text{O}},
\end{align}
where $r$ is the learning rate.

\subsection{Fine-tuning of the CLIP image encoder}
\label{subsec:train}
In this section, we fine-tune the CLIP image encoder by utilizing our dataset $\mathcal{D}^{pro}$. Specifically, after completing the generation process, we acquire a set of synthetic OOD images, denoted as $\mathcal{X}^\text{syn}$, which is paired with corresponding InD data. Each image in $\mathcal{X}^{syn}$ is fed into the CLIP model along with the negative label set $\mathcal{Y}^-$, aligning a negative label to each synthetic image and forming the dataset $\mathcal{D}^\text{syn}=\{(x^\text{syn}, y^-)\}, x^\text{syn}\in\mathcal{X}^\text{syn},y^-\in\mathcal{Y}_*^-$, where $\mathcal{Y}_*^-$ is the subset of negative labels associated with these images. Each generated OOD image is paired one-to-one with a corresponding InD image.

Using both InD data and these synthetic OOD samples, we create a training dataset $\mathcal{D}^{pro}=\mathcal{D}^{syn} \cup \mathcal{D}^\text{in}_*=\{(x_i, y_i)\}^{2m}_{i=1}$ for the image encoder fine-tuning, where $\mathcal{D}^\text{in}_*$ represents a subset of $\mathcal{D}^\text{in}$, and $m$ stands for the number of InD data we selected from $\mathcal{D}^{in}$. The selection of InD data is critical, as it directly affects the fine-tuning outcomes of the CLIP image encoder. To identify the most information-rich images within each category, we calculated the ratio of JPEG file size to the number of pixels for all images, sorted them accordingly, and then selected a specified number of top-ranked images from each category. This strategy ensures that we choose a batch of images with the highest complexity in each category.

As Fig.~\ref{fig:framework}(b) shows, The parameters of the CLIP image encoder $F$ remain frozen during training, and only the parameters of the projection layer $\delta$ are updated. We employ the CLIP loss $\mathcal{L}^\text{P}$ to train $\delta$:
\begin{align}
    \hat I_i = \delta(F(x_i)), 
    T_i = H(y^-_i), \quad (x_i, y_i)\in\mathcal{D}^{pro},\\
    \mathcal{L}^\text{P} = -\frac{1}{2m}\sum^{2m}_{i=1}\text{log}\frac{\text{exp}(sim(\hat I_i, T_i)/\tau)}{\sum^{M'}_{j=1}\text{exp}(sim(\hat I_i, T_j)/\tau)},
\end{align}
where $0<M'\leq M$ represents the number of negative labels in the subset, $H$ stands for CLIP text encoder,  $\hat I_i$ and $T_i$ stand for image feature and text feature, respectively, and $\tau$ is the temperature parameter.

\begin{table*}[t]
\caption{Comparison of OOD detection performance between SynOOD and existing methods. The best and second-best results are highlighted in \textbf{bold} and \underline{underlined}, respectively. All methods use ViT/B-16 as the backbone. Methods in the upper section are pre-trained on ImageNet, while those in the lower section utilize CLIP pre-training.}
\label{tab:main_results}
\footnotesize
\centering
\tabcolsep=0.17cm
\begin{tabular}{@{}lcccccccccccccc@{}}

\toprule
\multicolumn{1}{c}{}                                  & \multicolumn{12}{c}{\textbf{OOD Datasets}}                                                                                                                                                                                                           & \multicolumn{2}{c}{}                                   \\ \cmidrule(lr){2-13}
\multicolumn{1}{c}{}                                  & \multicolumn{2}{c}{\textbf{iNatrualist}} & \multicolumn{1}{l}{} & \multicolumn{2}{c}{\textbf{SUN}}  & \multicolumn{1}{l}{} & \multicolumn{2}{c}{\textbf{Place}} & \multicolumn{1}{l}{} & \multicolumn{2}{c}{\textbf{Texture}} & \multicolumn{1}{l}{} & \multicolumn{2}{c}{\multirow{-2}{*}{\textbf{Average}}} \\ \cmidrule(lr){2-3} \cmidrule(lr){5-6} \cmidrule(lr){8-9} \cmidrule(lr){11-12} \cmidrule(l){14-15} 
\multicolumn{1}{c}{\multirow{-3}{*}{\textbf{Method}}} & \textbf{AUROC↑}     & \textbf{FPR95↓}    &                      & \textbf{AUROC↑} & \textbf{FPR95↓} &                      & \textbf{AUROC↑}  & \textbf{FPR95↓} &                      & \textbf{AUROC↑}   & \textbf{FPR95↓}  &                      & \textbf{AUROC↑}            & \textbf{FPR95↓}                                                                           \\ \midrule
MSP~\cite{hendrycks2016baseline}                                                  & 87.44               & 58.36              &                      & 79.73           & 73.72           &                      & 79.67            & 74.41           &                      & 79.69             & 71.93            &                      & 81.63                      & 69.61                     \\
ODIN~\cite{liang2017enhancing}                                                  & 94.65               & 30.22              &                      & 87.17           & 54.04           &                      & 85.54            & 55.06           &                      & 87.85             & 51.67            &                      & 88.80                      & 47.75                     \\
Energy~\cite{liu2020energy}                                                & 95.33               & 26.12              &                      & 92.66           & 35.97           &                      & 91.41            & 39.87           &                      & 86.76             & 57.61            &                      & 91.54                      & 39.89                     \\
GradNorm~\cite{huang2021importance}                                              & 72.56               & 81.50              &                      & 72.86           & 82.00           &                      & 73.70            & 80.41           &                      & 70.26             & 79.36            &                      & 72.35                      & 80.82                     \\
ViM~\cite{wang2022vim}                                                   & 93.16               & 32.19              &                      & 87.19           & 54.01           &                      & 83.75            & 60.67           &                      & 87.18             & 53.94            &                      & 87.82                      & 50.20                     \\
KNN~\cite{sun2022out}                                                   & 94.52               & 29.17              &                      & 92.67           & 35.62           &                      & 91.02            & 39.61           &                      & 85.67             & 64.35            &                      & 90.97                      & 42.19                     \\
VOS~\cite{du2022vos}                                                   & 94.62               & 28.99              &                      & 92.57           & 36.88           &                      & 91.23            & 38.39           &                      & 86.33             & 61.02            &                      & 91.19                      & 41.32                     \\
DICE~\cite{sun2022dice}                                                  & 94.49               & 25.63              &                      & 90.83           & 35.15           &                      & 87.48            & 46.49           &                      & 90.30             & 31.72            &                      & 90.78                      & 34.75                     \\
ReAct~\cite{sun2021react}                                                 & 96.22               & 20.38              &                      & 94.20           & 24.20           &                      & 91.58            & 33.85           &                      & 89.80             & 47.30            &                      & 92.95                      & 31.43                     \\ \midrule
ZOC~\cite{esmaeilpour2022zero}                                                   & 86.09               & 87.30              &                      & 81.20           & 81.51           &                      & 83.39            & 73.06           &                      & 76.46             & 98.90            &                      & 81.79                      & 85.19                     \\
MCM~\cite{ming2022delving}                                                   & 94.59               & 32.20              &                      & 92.25           & 38.80           &                      & 90.31            & 46.20           &                      & 86.12             & 58.50            &                      & 90.82                      & 43.93                     \\
CoOp~\cite{zhou2022learning}                                                  & 94.89               & 29.47              &                      & 93.36           & 31.34           &                      & 90.07            & 40.28           &                      & 87.58             & 54.25            &                      & 91.48                      & 38.84                     \\
CoCoOp~\cite{zhou2022conditional}                                                & 94.73               & 30.74              &                      & 93.15           & 31.18           &                      & 90.63            & 38.75           &                      & 87.92             & 53.84            &                      & 91.61                      & 38.63                     \\
NPOS~\cite{tao2023non}                                                  & 96.19               & 16.58              &                      & 90.44           & 43.77           &                      & 89.44            & 45.27           &                      & 88.80             & 46.12            &                      & 91.22                      & 37.94                     \\
CLIPN~\cite{wang2023clipn}                                                 & 95.27               & 23.94              &                      & 93.93           & 26.17           &                      &  92.28      & 33.45     &                      & 90.93       & 40.83            &                      & 93.10                      & 31.10                     \\
LSN~\cite{nie2024out}                                                   & 95.83               & 21.56              &                      & 94.35           & 26.32           &                      & 91.25            & 34.48           &                      & 90.42             & 38.54      &                      & 92.96                      & 30.23                     \\
LoCoOp~\cite{miyai2024locoop}                                                & 96.86               & 16.05              &                      & 95.07           & 23.44           &                      & 91.98            & 32.87           &                      & 90.19             & 42.18            &                      & 93.53                      & 28.64                     \\
NegLabel~\cite{jiang2024negative}                                              & 99.49         & 1.91         & {\ul }               & 95.49     & 20.53     &                      & 91.64            & 35.59           &                      & 90.22             & 43.56            &                      &  94.21                & 25.40               \\
CSP\cite{chen2024conjugated}                                                   & {\ul 99.60}         & {\ul 1.54}         &           & {\ul 96.66}     & {\ul 13.66}     & \multicolumn{1}{l}{} & 92.90            & {\ul 29.32}     & \multicolumn{1}{l}{} & 93.86             & {\ul 25.52}      & \multicolumn{1}{l}{} & 95.76                      & {\ul 17.51}                     \\
AdaNeg\cite{zhang2024adaneg}                                                & \textbf{99.71}      & \textbf{0.59}      &           & \textbf{97.44}  & \textbf{9.50}   &                      & {\ul 94.55}      & 34.34           &                      & {\ul 94.93}       & 31.27            &                      & {\ul 96.66}                      & 18.92                     \\

\rowcolor[HTML]{F2F2F2} 
\textbf{SynOOD (Ours)}                                         & 99.57      & 1.57      & \textbf{}            & 95.82  & 20.46  & \textbf{}            & \textbf{97.37}   & \textbf{12.12}  & \textbf{}            & \textbf{95.29}    & \textbf{22.94}   & \textbf{}            & \textbf{97.01}             & \textbf{14.27}            \\ \bottomrule
\end{tabular}
\vspace{-1em}
\end{table*}

\subsection{Fine-tuning of the CLIP text encoder features}
\label{subsec:finetune}
The primary motivation for fine-tuning negative label features derived from the CLIP text encoder $H$ is to ensure that the semantic representations of negative labels adapt specifically to OOD data with subtle variations from InD. While the fine-tuned CLIP image encoder aligns images to the corresponding negative labels, it may not fully capture the nuances of the synthetic OOD samples without some adaptation on the text side. This fine-tuning dynamically adjusts these representations for better generalization and reduces overfitting risks of the image encoder fine-tuning on a limited set of synthetic OOD data. 

Specifically, we utilize CLIP and make a subset of the negative label, $\mathcal{Y}_*^-$, associated with synthetic OOD samples $\mathcal{D}^{\text{syn}}$, learnable. We denote the CLIP text features of $\mathcal{Y}_*^-$ as $\mathcal{T}_*^{neg}=\{T^{neg}_i\}^{M'}_{i=1}$, where $M'\approx\frac{1}{2}M$, leaving the remaining labels in $\mathcal{Y}^-$ fixed. This subset negative label fine-tuning reduces the semantic gap between InD and negative labels while maintaining model robustness, enabling precise detection of near-boundary OOD samples.

During fine-tuning, the learnable features $\mathcal{T}_*^{neg}$ are adjusted based on synthetic OOD images $\mathcal{X}^\text{syn}$, allowing the negative label features to move closer in feature space to these OOD samples while maintaining separation from InD representations. The objective is to align negative labels to capture relevant distinctions from InD data without overlap. This direct fine-tuning approach reduces the computational cost of modifying text encoder embeddings. The loss function $\mathcal{L}^\text{T}$ for fine-tuning negative features derived from text encoder $H$ is:
\begin{align}
    I^\text{syn}_i &= F(x^\text{syn}_i), \\
    \mathcal{L}^\text{T} &= -\frac{1}{m}\sum^{m}_{i=1}\text{log}\frac{\text{exp}(\text{sim}(I_i, T^\text{neg}_i)/\tau)}{\sum^{M'}_{j=1}\text{exp}(\text{sim}( T^\text{neg}_i, I_j)/\tau)}.
\end{align}
Through fine-tuning, the negative text encoder features are adjusted better to capture the distinctions between InD and OOD data. We do not use the fine-tuned CLIP image encoder, as we aim to avoid adapting the image feature projection based on the text features. This approach helps prevent the negative label features from shifting to a suboptimal position, ensuring better control over their alignment. We will provide a more detailed discussion in the experiments section.

\section{Experiments}
\subsection{Experimental Setup}
\textbf{Dataset.} We follow Huang et al.~\cite{huang2021mos} and conduct extensive experiments with the standard large-scale ImageNet-1k~\cite{russakovsky2015imagenet} as InD data. For OOD data, we employ iNaturalist~\cite{van2018inaturalist}, SUN~\cite{xiao2010sun}, Places365~\cite{zhou2017places}, and Texture~\cite{cimpoi2014describing}. Moreover, we test our SynOOD on OpenOOD benchmark~\cite{yang2022openood, zhang2023openood}. Specifically, ImageNet-O~\cite{hendrycks2021natural}, SSB-hard~\cite{vaze2021open} and NINCO~\cite{bitterwolf2023or} are labeled as near-OOD, and far-OOD contains iNaturalist~\cite{van2018inaturalist}, Texture~\cite{cimpoi2014describing}, and OpenImage-O~\cite{wang2022vim}.

\noindent\textbf{Computational Cost.} Compared to NegLabel, our method adds less than 1\% additional parameters and takes under 2 ms per image during inference.

\noindent\textbf{Implementation Details.} We use LLaVA~\cite{liu2024visual} to generate prompts for the diffusion model and employ Stable Diffusion 2 for inpainting~\cite{rombach2022high} as the generative model to create near-boundary OOD images. We set the strength parameter to 0.6 and the number of timesteps to 20. Energy~\cite{liu2020energy} is utilized as the OOD Loss function, with ResNet50 serving as the backbone model. The inpainting process iterates 3 times, with $r\cdot\rho=10$. For the CLIP image encoder fine-tuning, we only train for 3 epochs using Adam with a learning rate of $1\times 10^{-3}$, a batch size of 128, and a weight decay of $1\times10^{-5}$. For the CLIP text encoder features fine-tuning, we employ SGD with a learning rate of $2\times10^{-3}$ and train for 5 epochs. All experiments are performed using PyTorch~\cite{paszke2019pytorch} on two NVIDIA V100.

\subsection{Main Results}
 As presented in Table~\ref{tab:main_results}, we evaluate SynOOD against a range of existing OOD detection approaches on the widely-used ImageNet-1k benchmark, showcasing its performance across multiple challenging datasets. The methods listed from MSP~\cite{hendrycks2016baseline} to ReAct~\cite{sun2021react} represent OOD detection approaches based on single-modal vision networks, while the methods from ZOC~\cite{esmaeilpour2022zero} to NegLabel~\cite{jiang2024negative} employ the multi-modal capabilities of the CLIP model. The results consistently demonstrate that pre-trained multi-modal models like CLIP have a significant advantage over traditional single-modal vision networks for OOD detection, underscoring the benefits of aligning both text and visual representations to improve OOD performance. When comparing SynOOD to NegLabel, we observe that SynOOD maintains strong detection performance on the iNaturalist~\cite{van2018inaturalist} and SUN~\cite{xiao2010sun} datasets, which involve natural images and complex scenes, respectively. More notably, SynOOD achieves significant improvements on the Places~\cite{zhou2017places} and Texture~\cite{cimpoi2014describing} datasets, which feature a broader diversity of environmental and textural variations. These improvements underscore SynOOD’s ability to more accurately capture and represent OOD boundaries in data with high intra-class variability and complex visual patterns, areas where traditional methods often struggle.
Overall, SynOOD establishes a new state-of-the-art in OOD detection, surpassing previous methods with a substantial AUROC improvement of 2.80\% and an FPR95 reduction of 11.13\%. This performance boost reflects SynOOD’s robust design and effective use of negative label fine-tuning and iterative OOD sample generation, which together enable a more nuanced alignment between InD and near-boundary OOD samples.

\begin{table}[t]
\footnotesize
\centering
\tabcolsep=0.10cm
\caption{OOD detection performance on the OpenOOD benchmark. The methods in the upper section are using the whole ImageNet for training.  Results are averaged across OOD datasets.}
\label{tab:nearfar}
\begin{tabular}{@{}l|cc|cc@{}}
\toprule
\multicolumn{1}{c|}{\multirow{2}{*}{\textbf{Method}}} & \multicolumn{2}{c|}{\textbf{AUROC↑}} & \multicolumn{2}{c}{\textbf{FPR95↓}} \\ \cmidrule(l){2-5} 
\multicolumn{1}{c|}{}                                 & \textbf{NearOOD}  & \textbf{FarOOD} & \textbf{NearOOD} & \textbf{FarOOD} \\ \midrule
GEN~\cite{liu2023gen}                                 & 78.97              & 90.98           & -                 & -               \\
AugMix~\cite{hendrycks2019augmix}+ReAct~\cite{sun2021react} & 79.94        & 93.70           & -                 & -               \\
RMDS~\cite{ren2021simple}                               & 80.09              & 92.60           & -                 & -               \\
AugMix~\cite{hendrycks2019augmix}+ASH~\cite{djurisic2022extremely}                                           & 82.16              & 96.05           & -                 & -               \\ \midrule
MCM~\cite{ming2022delving}                                         & 59.89              & 80.71           & 81.02             & 68.88           \\
NegLabel~\cite{jiang2024negative}                           & 75.47              & 94.30           & 74.74             & 25.73           \\
\rowcolor[HTML]{F2F2F2}\textbf{Ours}                                         & \textbf{77.55}     & \textbf{96.21}  & \textbf{71.68}    & \textbf{17.11}  \\ \bottomrule
\end{tabular}
\vspace{-1em}
\end{table}

\begin{table*}[t]
\footnotesize
\tabcolsep=0.14cm
\caption{OOD detection performance comparison across various CLIP architectures. Results are averaged across four OOD datasets.}
\label{tab:backbone}
\begin{tabular}{@{}ccccccclcclcclcc@{}}
\toprule
                                    &                                         & \multicolumn{12}{c}{\textbf{OOD Datasets}}                                                                                                                                                                                                                                                                                                                                                                                                                                                                                                       & \multicolumn{2}{c}{}                                                            \\ \cmidrule(lr){3-14}
                                    &                                         & \multicolumn{2}{c}{\textbf{iNatrualist}}                                       & \textbf{}                         & \multicolumn{2}{c}{\textbf{SUN}}                                                &                                                       & \multicolumn{2}{c}{\textbf{Place}}                                              &                                                       & \multicolumn{2}{c}{\textbf{Texture}}                                            &                                                       & \multicolumn{2}{c}{\multirow{-2}{*}{\textbf{Average}}}                          \\ \cmidrule(lr){3-4} \cmidrule(lr){6-7} \cmidrule(lr){9-10} \cmidrule(lr){12-13} \cmidrule(l){15-16} 
\multirow{-3}{*}{\textbf{Backbone}} & \multirow{-3}{*}{\textbf{Method}}       & \textbf{AUROC↑}                        & \textbf{FPR95↓}                       &                                   & \textbf{AUROC↑}                        & \textbf{FPR95↓}                        & \multicolumn{1}{c}{}                                  & \textbf{AUROC↑}                        & \textbf{FPR95↓}                        & \multicolumn{1}{c}{}                                  & \textbf{AUROC↑}                        & \textbf{FPR95↓}                        & \multicolumn{1}{c}{}                                  & \textbf{AUROC↑}                        & \textbf{FPR95↓}                        \\ \midrule
                                    & NegLabel                                & \textbf{99.24}                         & \textbf{2.88}                         &                                   & 94.54                                  & 26.51                                  &                                                       & 89.72                                  & 42.60                                   &                                                       & 88.40                                   & 50.8                                   &                                                       & 92.97                                  & 30.70                                   \\
\multirow{-2}{*}{ResNet50}          & \cellcolor[HTML]{F2F2F2}SynOOD & \cellcolor[HTML]{F2F2F2}98.89 & \cellcolor[HTML]{F2F2F2}4.26 & \cellcolor[HTML]{F2F2F2}\textbf{} & \cellcolor[HTML]{F2F2F2}\textbf{95.13} & \cellcolor[HTML]{F2F2F2}\textbf{24.98} & \cellcolor[HTML]{F2F2F2}                              & \cellcolor[HTML]{F2F2F2}\textbf{96.07} & \cellcolor[HTML]{F2F2F2}\textbf{16.29} & \cellcolor[HTML]{F2F2F2}                              & \cellcolor[HTML]{F2F2F2}\textbf{93.09} & \cellcolor[HTML]{F2F2F2}\textbf{34.65} & \cellcolor[HTML]{F2F2F2}                              & \cellcolor[HTML]{F2F2F2}\textbf{95.80} & \cellcolor[HTML]{F2F2F2}\textbf{20.05} \\ \midrule
                                    & NegLabel                                & 99.11                                  & 3.73                                  &                                   & \textbf{95.27}                         & \textbf{22.48}                         &                                                       & 91.92                                  & 34.94                                  &                                                       & 88.57                                  & 50.51                                  &                                                       & 93.67                                  & 27.92                                  \\
\multirow{-2}{*}{ViT-B/32}          & \cellcolor[HTML]{F2F2F2}SynOOD          & \cellcolor[HTML]{F2F2F2}\textbf{99.35} & \cellcolor[HTML]{F2F2F2}\textbf{2.42} & \cellcolor[HTML]{F2F2F2}          & \cellcolor[HTML]{F2F2F2}94.90           & \cellcolor[HTML]{F2F2F2}25.10           & \cellcolor[HTML]{F2F2F2}                              & \cellcolor[HTML]{F2F2F2}\textbf{97.20}  & \cellcolor[HTML]{F2F2F2}\textbf{13.24} & \cellcolor[HTML]{F2F2F2}                              & \cellcolor[HTML]{F2F2F2}\textbf{92.26} & \cellcolor[HTML]{F2F2F2}\textbf{36.86} & \cellcolor[HTML]{F2F2F2}                              & \cellcolor[HTML]{F2F2F2}\textbf{95.93} & \cellcolor[HTML]{F2F2F2}\textbf{19.41} \\ \midrule
                                    & NegLabel                                & 99.49                                  & 1.91                                  &                                   & 95.49                                  & 20.53                                  & \multicolumn{1}{c}{}                                  & 91.64                                  & 35.59                                  & \multicolumn{1}{c}{}                                  & 90.22                                  & 43.56                                  & \multicolumn{1}{c}{}                                  & 94.21                                  & 25.40                                  \\
\multirow{-2}{*}{ViT-B/16}          & \cellcolor[HTML]{F2F2F2}SynOOD          & \cellcolor[HTML]{F2F2F2}\textbf{99.57} & \cellcolor[HTML]{F2F2F2}\textbf{1.57} & \cellcolor[HTML]{F2F2F2}\textbf{} & \cellcolor[HTML]{F2F2F2}\textbf{95.82} & \cellcolor[HTML]{F2F2F2}\textbf{20.46} & \multicolumn{1}{c}{\cellcolor[HTML]{F2F2F2}\textbf{}} & \cellcolor[HTML]{F2F2F2}\textbf{97.37} & \cellcolor[HTML]{F2F2F2}\textbf{12.12} & \multicolumn{1}{c}{\cellcolor[HTML]{F2F2F2}\textbf{}} & \cellcolor[HTML]{F2F2F2}\textbf{95.29} & \cellcolor[HTML]{F2F2F2}\textbf{22.94} & \multicolumn{1}{c}{\cellcolor[HTML]{F2F2F2}\textbf{}} & \cellcolor[HTML]{F2F2F2}\textbf{97.01} & \cellcolor[HTML]{F2F2F2}\textbf{14.27} \\ \bottomrule
\end{tabular}
\vspace{-1em}
\end{table*}

\noindent\textbf{Evaluation on OpenOOD benchmark.} We further evaluate SynOOD on the OpenOOD benchmark, which includes both near-OOD and far-OOD scenarios. As presented in Table~\ref{tab:nearfar}, the methods in the upper section are drawn from OpenOOD~\cite{zhang2023openood}. These methods typically show stronger performance in near-OOD detection, as they benefit from training on the full ImageNet dataset, which contains over 1.2 million images, giving them a substantial advantage. This allows them to capture diverse InD patterns, improving near-OOD detection accuracy. In contrast, SynOOD uses only a lightweight subset of 50k ImageNet images yet achieves competitive performance, outperforming all methods in far-OOD detection and exceeding MCM~\cite{ming2022delving} and NegLabel~\cite{jiang2024negative} on near-OOD detection. These findings underscore SynOOD’s effectiveness across both near and far-OOD conditions, even with limited training data. This balance highlights the robustness of our approach, particularly in the challenging far-OOD scenario, where our method consistently maintains superior discrimination. These results confirm the generalization capability of SynOOD and its adaptability across various OOD conditions.

\subsection{Ablation Study}
\textbf{Image Generation and Training Components.}  In Tab.~\ref{tab:ablation}, we investigate the effect of the fine-tuning of the CLIP image encoder and the fine-tuning of the CLIP text encoder features using various synthetic image generation strategies. We compare two image generation approaches for synthesizing OOD data: (1) directly generating images using negative labels as prompts in a text-to-image diffusion model and (2) using an iterative image generation process that refines the alignment between generated images and their associated negative labels. The first method, direct generation, employs negative labels as prompts to synthesize images in a single pass through a diffusion model. While effective, this approach can sometimes yield images with limited similarity with InD data, which may not fully capture the nuanced distinctions between InD and near-boundary OOD samples. The iterative approach substantially improves performance across both image encoder fine-tuning and label feature fine-tuning, as it progressively shifts the image away from the original theme while still preserving visual connections to the InD data through the background features. Furthermore, our experiments indicate that the best average performance is achieved when both the image encoder fine-tuning and negative label fine-tuning are jointly applied with the iterative generation strategy.

\begin{table}[t]
\footnotesize
\centering
\tabcolsep=0.10cm
\caption{A set of ablation experiments on SynOOD. \textbf{“FT Label”} refers to fine-tuning the label features, \textbf{“Neg Image”} indicates images generated by a text-to-image diffusion model prompted by the negative labels, and \textbf{“Grad Image”} represents images produced using our iterative generation process. Results are averaged across four OOD datasets.}
\label{tab:ablation}
\begin{tabular}{@{}cc|cc|cc@{}}
\toprule
                                      &                                              & \multicolumn{2}{c|}{\textbf{Data}}                            & \multicolumn{2}{c}{\textbf{Average}} \\ \cmidrule(l){3-6} 
\multirow{-2}{*}{\textbf{Projection}} & \multirow{-2}{*}{\textbf{FT Label}} & \textbf{Neg Image} & \textbf{Grad Image} & \textbf{AUROC↑}   & \textbf{FPR95↓}  \\ \midrule
-                                     & -                                            & -                             & -                             & 94.21             & 25.40            \\
\checkmark                                     &                                              & \checkmark                             &                               & 94.85             & 22.59            \\
\checkmark                                     &                                              &                               & \checkmark                             & {\ul 96.21}       & {\ul 17.29}      \\
\checkmark                                     & \checkmark                                            & \checkmark                             &                               & 95.01             & 21.93            \\
\rowcolor[HTML]{F2F2F2} 
\checkmark                                     & \checkmark                                            &                               & \checkmark                             & \textbf{97.01}    & \textbf{14.27}   \\ \bottomrule
\end{tabular}
\vspace{-1em}
\end{table}

% \noindent\textbf{Effect of the number of synthetic data.} Figure~\ref{fig:numofdata} illustrates the impact of varying amounts of synthetic OOD data on SynOOD when training both the CLIP image encoder fine-tuning and negative text encoder features fine-tuning. Specifically, we evaluate SynOOD with the number of synthetic OOD samples $m$ set to $\{1k, 5k, 10k, 20k, 30k, 50k, 75k, 100k\}$. Overall, SynOOD demonstrates stable performance, with accuracy improving as $m$ increases, underscoring the effectiveness of our iterative data generation strategy in enhancing OOD detection.
% However, when $m$ surpasses 50k, a slight decline in performance becomes evident. This decrease can be attributed to the balance between fixed and fine-tuned negative label features. Our method fine-tunes approximately 5.5k out of 11k negative labels, preserving SynOOD’s robust performance on far-boundary OOD data while increasing sensitivity to near-boundary samples. However, fine-tuning over 7k labels (e.g., with $m$ set to 75k) disrupts this balance; excessive fine-tuning of text features can misclassify easier OOD samples, causing a slight performance decline.
% Furthermore, the results reveal that the inclusion of fine-tuned negative label features consistently boosts performance across all $m$ values compared to training without fine-tuning. This benefit is observed irrespective of the amount of synthetic data, highlighting the significance of tailored feature alignment in maintaining effective OOD detection boundaries. 

\begin{figure}[t]
    \centering
    \includegraphics[width=1\linewidth]{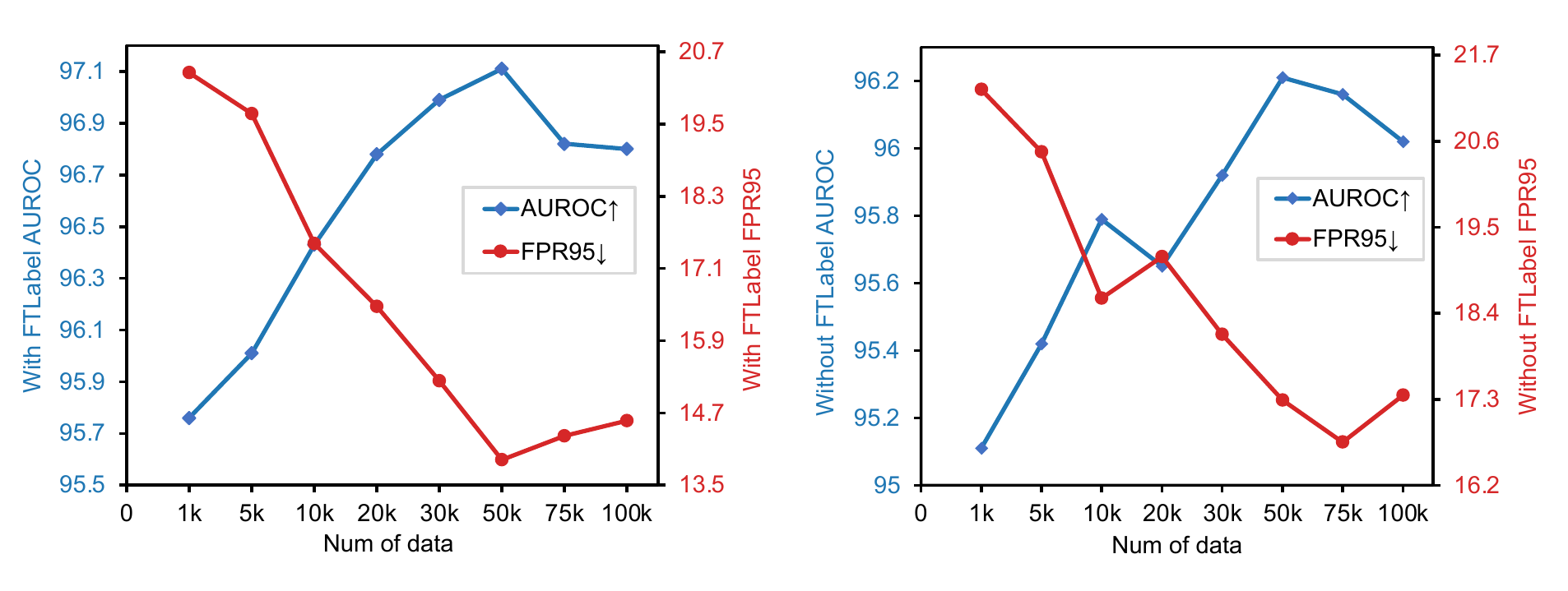}
    \caption{SynOOD performance with different amounts of synthetic OOD data. The left plot shows results with fine-tuning negative label features, while the right plot shows results without fine-tuning. Results are averaged across four OOD datasets.}
    \label{fig:numofdata}
    \vspace{-1em}
\end{figure}

\noindent\textbf{Effect of the number of synthetic data.} Figure~\ref{fig:numofdata} shows how varying the amount of synthetic OOD data affects SynOOD when fine-tuning both the CLIP image encoder and negative text encoder features. We evaluate SynOOD with $m \in \{1k, 5k, 10k, 20k, 30k, 50k, 75k, 100k\}$ synthetic OOD samples. SynOOD maintains stable performance, with accuracy improving as $m$ increases, demonstrating the effectiveness of our iterative data generation strategy. However, performance slightly declines when $m$ exceeds 50k, due to the balance between fixed and fine-tuned negative label features. Our method fine-tunes about 5.5k of 11k negative labels, preserving strong performance on far-boundary OOD data and enhancing sensitivity to near-boundary samples. Fine-tuning over 7k labels (e.g., $m = 75k$) disrupts this balance, leading to misclassification of easier OOD samples and a minor performance drop. Additionally, including fine-tuned negative label features consistently improves performance across all $m$ values compared to training without fine-tuning, underscoring the importance of tailored feature alignment for effective OOD detection.

\begin{figure*}[t]
    \centering
    \includegraphics[width=1\linewidth]{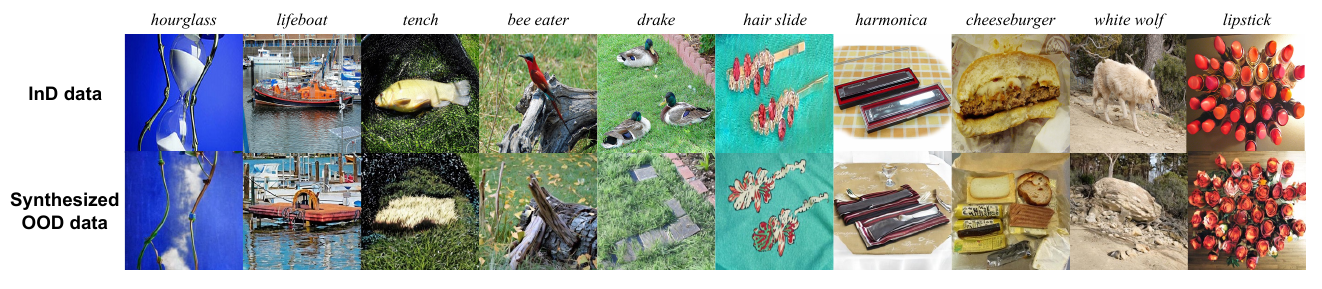}
    \caption{ Visualization of InD and synthetic OOD data. The labels at the top of the figure represent ImageNet categories as InD. The first row shows ImageNet (InD) images, while the second row presents our synthetic OOD data.}
    \label{fig:visualize}
\vspace{-0.5em}
\end{figure*}

\noindent\textbf{Anaylsis of different CLIP networks.} In Table~\ref{tab:backbone}, we evaluate the effectiveness of our method, SynOOD, across a range of CLIP-based architectures, including ResNet50~\cite{he2016deep}, ViT-B/32~\cite{dosovitskiy2021imageworth16x16words}, and ViT-B/16~\cite{dosovitskiy2021imageworth16x16words}. SynOOD consistently outperforms the baseline, NegLabel~\cite{jiang2024negative}, demonstrating superior performance across all architectures. Notably, our method yields significant improvements in the FPR95 metric, achieving reductions exceeding 10\% on each network, which is a substantial enhancement for high-confidence OOD detection. This performance boost indicates that SynOOD is not only effective in lowering false positive rates but also exhibits strong robustness and adaptability across diverse backbone architectures. The consistent gains achieved across both convolutional (ResNet) and transformer-based (ViT) models underscore the generalizability of our approach, showing that SynOOD’s design principles are broadly applicable to various network structures within the CLIP framework. This adaptability further emphasizes SynOOD’s potential for application in a wide range of OOD detection scenarios.

\begin{table}[t]
\centering
\footnotesize
\tabcolsep=0.12cm
\caption{Performance comparison across training strategies, including joint training and step-by-step training with or without the trained projection layer during fine-tuning. Results are averaged across four OOD datasets.}
\label{tab:strategy}
\begin{tabular}{@{}l|c|cc@{}}
\toprule
\multicolumn{1}{c|}{\multirow{2}{*}{\textbf{Strategy}}} & \multirow{2}{*}{\textbf{Fine-Tuning with Projection}} & \multicolumn{2}{c}{\textbf{Average}} \\ \cmidrule(l){3-4} 
\multicolumn{1}{c|}{}                                     &                                                & \textbf{AUROC↑}   & \textbf{FPR95↓}  \\ \midrule
Joint                                                     & -                                              & 95.93             & 19.41            \\
Separate                                                & \Checkmark                                           & 96.49             & 16.18            \\
Separate                                                & \XSolidBrush                                              & \textbf{97.01}    & \textbf{14.27}   \\ \bottomrule
\end{tabular}
\vspace{-1.5em}
\end{table}

\noindent\textbf{Training Strategy Comparison.} In Table~\ref{tab:strategy}, we examine three distinct training strategies to evaluate our SynOOD framework: joint training and two step-by-step training approaches. For joint training, we optimize both the projection layer and label feature fine-tuning. The step-by-step approach separates the training into sequential phases. Specifically, we implement two variations of step-by-step training: one that incorporates the pre-trained projection layer during the label feature fine-tuning phase, and another that excludes it. The results in Table~\ref{tab:strategy} reveal that step-by-step training is more effective than joint training, providing both enhanced stability during training and improved detection performance. This improvement is due to the sequential focus on each component, which may reduce interference effects seen in joint training. Notably, our experiments indicate that fine-tuning the label features without image encoder fine-tuning yields the best results. We hypothesize that the projection layer, trained on only 50k synthetic OOD samples, is prone to overfitting, which may limit generalization when combined with fine-tuned label features. By excluding the projection layer in this phase, SynOOD achieves a better balance between specificity and robustness in OOD detection. This evaluation of training strategies underscores the importance of carefully structuring the training process for complex OOD detection systems.

\noindent\textbf{Visualize of the synthetic Data.} In Fig.~\ref{fig:visualize}, we present several InD and synthetic OOD data pairs to illustrate how our method generates OOD samples that are visually similar to InD samples, yet exhibit clear OOD characteristics. These examples show that the synthetic OOD images closely resemble their corresponding InD images to the human eye, but contain subtle differences that distinguish them as OOD. For instance, in the hourglass image on the left of Fig.~\ref{fig:visualize}, the original InD sample depicts an hourglass with white sand, slim supports, and a blue background. Our generation process has transformed this scene into an image with a blue sky, white clouds, and vines, making it perceptually similar yet meaningfully different from the InD data. Similarly, on the right, lipsticks are reimagined as flowers in the same setting. Our method can even decompose a cheeseburger, displaying each component separately against a consistent background. We hope our iterative generation process inspires further research into innovative OOD data generation techniques and opens new possibilities for other applications where similar approaches might be beneficial.

\section{Conclusion}
In this paper, we present SynOOD, a novel approach to OOD detection that combines iterative generative techniques with fine-tuned CLIP models and features for enhanced identification of challenging OOD samples. By synthesizing near-boundary OOD samples using a diffusion model, SynOOD generates data that is visually similar to, yet semantically distinct from, InD data, allowing for more precise OOD discrimination. Extensive evaluations on multiple benchmark datasets demonstrate that SynOOD surpasses existing methods, achieving state-of-the-art performance in AUROC and FPR95 metrics. Our work highlights the effectiveness of synthetic data for OOD detection, suggesting new directions for using generative methods to improve model robustness in diverse tasks.  We believe SynOOD opens up new directions for OOD detection and encourages future research into similar generative strategies for improving model robustness across other tasks.

\section{Acknowledgement}
This work was supported by National Natural Science Foundation of China (No.62072112) and Shanghai Science and Technology Committee under Grant (No. 24511103900, 24511103202) and was partly supported by National Key RD Program of China under grant No. 2022YFC3601405. 

{
    \small
    \bibliographystyle{ieeenat_fullname}
    \bibliography{main}

\begin{thebibliography}{10}
\providecommand{\url}[1]{#1}
\csname url@samestyle\endcsname
\providecommand{\newblock}{\relax}
\providecommand{\bibinfo}[2]{#2}
\providecommand{\BIBentrySTDinterwordspacing}{\spaceskip=0pt\relax}
\providecommand{\BIBentryALTinterwordstretchfactor}{4}
\providecommand{\BIBentryALTinterwordspacing}{\spaceskip=\fontdimen2\font plus
\BIBentryALTinterwordstretchfactor\fontdimen3\font minus \fontdimen4\font\relax}
\providecommand{\BIBforeignlanguage}[2]{{%
\expandafter\ifx\csname l@#1\endcsname\relax
\typeout{** WARNING: IEEEtran.bst: No hyphenation pattern has been}%
\typeout{** loaded for the language `#1'. Using the pattern for}%
\typeout{** the default language instead.}%
\else
\language=\csname l@#1\endcsname
\fi
#2}}
\providecommand{\BIBdecl}{\relax}
\BIBdecl

\bibitem{hendrycks2016baseline}
D.~Hendrycks and K.~Gimpel, ``A baseline for detecting misclassified and out-of-distribution examples in neural networks,'' \emph{arXiv preprint arXiv:1610.02136}, 2016.

\bibitem{liang2017enhancing}
S.~Liang, Y.~Li, and R.~Srikant, ``Enhancing the reliability of out-of-distribution image detection in neural networks,'' \emph{arXiv preprint arXiv:1706.02690}, 2017.

\bibitem{liu2020energy}
W.~Liu, X.~Wang, J.~Owens, and Y.~Li, ``Energy-based out-of-distribution detection,'' \emph{Advances in neural information processing systems}, vol.~33, pp. 21\,464--21\,475, 2020.

\bibitem{huang2021importance}
R.~Huang, A.~Geng, and Y.~Li, ``On the importance of gradients for detecting distributional shifts in the wild,'' \emph{Advances in Neural Information Processing Systems}, vol.~34, pp. 677--689, 2021.

\bibitem{wang2022vim}
H.~Wang, Z.~Li, L.~Feng, and W.~Zhang, ``Vim: Out-of-distribution with virtual-logit matching,'' in \emph{Proceedings of the IEEE/CVF conference on computer vision and pattern recognition}, 2022, pp. 4921--4930.

\bibitem{sun2022out}
Y.~Sun, Y.~Ming, X.~Zhu, and Y.~Li, ``Out-of-distribution detection with deep nearest neighbors,'' in \emph{International Conference on Machine Learning}.\hskip 1em plus 0.5em minus 0.4em\relax PMLR, 2022, pp. 20\,827--20\,840.

\bibitem{du2022vos}
X.~Du, Z.~Wang, M.~Cai, and Y.~Li, ``Vos: Learning what you don't know by virtual outlier synthesis,'' \emph{arXiv preprint arXiv:2202.01197}, 2022.

\bibitem{sun2022dice}
Y.~Sun and Y.~Li, ``Dice: Leveraging sparsification for out-of-distribution detection,'' in \emph{European Conference on Computer Vision}.\hskip 1em plus 0.5em minus 0.4em\relax Springer, 2022, pp. 691--708.

\bibitem{sun2021react}
Y.~Sun, C.~Guo, and Y.~Li, ``React: Out-of-distribution detection with rectified activations,'' \emph{Advances in Neural Information Processing Systems}, vol.~34, pp. 144--157, 2021.

\bibitem{esmaeilpour2022zero}
S.~Esmaeilpour, B.~Liu, E.~Robertson, and L.~Shu, ``Zero-shot out-of-distribution detection based on the pre-trained model clip,'' in \emph{Proceedings of the AAAI conference on artificial intelligence}, vol.~36, no.~6, 2022, pp. 6568--6576.

\bibitem{ming2022delving}
Y.~Ming, Z.~Cai, J.~Gu, Y.~Sun, W.~Li, and Y.~Li, ``Delving into out-of-distribution detection with vision-language representations,'' \emph{Advances in neural information processing systems}, vol.~35, pp. 35\,087--35\,102, 2022.

\bibitem{zhou2022learning}
K.~Zhou, J.~Yang, C.~C. Loy, and Z.~Liu, ``Learning to prompt for vision-language models,'' \emph{International Journal of Computer Vision}, vol. 130, no.~9, pp. 2337--2348, 2022.

\bibitem{zhou2022conditional}
------, ``Conditional prompt learning for vision-language models,'' in \emph{Proceedings of the IEEE/CVF conference on computer vision and pattern recognition}, 2022, pp. 16\,816--16\,825.

\bibitem{tao2023non}
L.~Tao, X.~Du, X.~Zhu, and Y.~Li, ``Non-parametric outlier synthesis,'' \emph{arXiv preprint arXiv:2303.02966}, 2023.

\bibitem{wang2023clipn}
H.~Wang, Y.~Li, H.~Yao, and X.~Li, ``Clipn for zero-shot ood detection: Teaching clip to say no,'' in \emph{Proceedings of the IEEE/CVF International Conference on Computer Vision}, 2023, pp. 1802--1812.

\bibitem{nie2024out}
J.~Nie, Y.~Zhang, Z.~Fang, T.~Liu, B.~Han, and X.~Tian, ``Out-of-distribution detection with negative prompts,'' in \emph{The Twelfth International Conference on Learning Representations}, 2024.

\bibitem{miyai2024locoop}
A.~Miyai, Q.~Yu, G.~Irie, and K.~Aizawa, ``Locoop: Few-shot out-of-distribution detection via prompt learning,'' \emph{Advances in Neural Information Processing Systems}, vol.~36, 2024.

\bibitem{jiang2024negative}
X.~Jiang, F.~Liu, Z.~Fang, H.~Chen, T.~Liu, F.~Zheng, and B.~Han, ``Negative label guided ood detection with pretrained vision-language models,'' \emph{arXiv preprint arXiv:2403.20078}, 2024.

\bibitem{Li2023hvcm}
J.~Li, X.~Zhou, P.~Guo, Y.~Sun, Y.~Huang, W.~Ge, and W.~Zhang, ``Hierarchical visual categories modeling: A joint representation learning and density estimation framework for out-of-distribution detection,'' in \emph{Proceedings of the IEEE/CVF International Conference on Computer Vision (ICCV)}, October 2023, pp. 23\,425--23\,435.

\bibitem{li2024tagood}
J.~Li, X.~Zhou, K.~Jiang, L.~Hong, P.~Guo, Z.~Chen, W.~Ge, and W.~Zhang, ``Tagood: A novel approach to out-of-distribution detection via vision-language representations and class center learning,'' \emph{arXiv preprint arXiv:2408.15566}, 2024.

\bibitem{miller1995wordnet}
G.~A. Miller, ``Wordnet: a lexical database for english,'' \emph{Communications of the ACM}, vol.~38, no.~11, pp. 39--41, 1995.

\bibitem{radford2021learning}
A.~Radford, J.~W. Kim, C.~Hallacy, A.~Ramesh, G.~Goh, S.~Agarwal, G.~Sastry, A.~Askell, P.~Mishkin, J.~Clark \emph{et~al.}, ``Learning transferable visual models from natural language supervision,'' in \emph{International conference on machine learning}.\hskip 1em plus 0.5em minus 0.4em\relax PMLR, 2021, pp. 8748--8763.

\bibitem{liu2023llava}
H.~Liu, C.~Li, Q.~Wu, and Y.~J. Lee, ``Visual instruction tuning,'' 2023.

\bibitem{qwen}
J.~Bai, S.~Bai, Y.~Chu, Z.~Cui, K.~Dang, X.~Deng, Y.~Fan, W.~Ge, Y.~Han, F.~Huang, B.~Hui, L.~Ji, M.~Li, J.~Lin, R.~Lin, D.~Liu, G.~Liu, C.~Lu, K.~Lu, J.~Ma, R.~Men, X.~Ren, X.~Ren, C.~Tan, S.~Tan, J.~Tu, P.~Wang, S.~Wang, W.~Wang, S.~Wu, B.~Xu, J.~Xu, A.~Yang, H.~Yang, J.~Yang, S.~Yang, Y.~Yao, B.~Yu, H.~Yuan, Z.~Yuan, J.~Zhang, X.~Zhang, Y.~Zhang, Z.~Zhang, C.~Zhou, J.~Zhou, X.~Zhou, and T.~Zhu, ``Qwen technical report,'' \emph{arXiv preprint arXiv:2309.16609}, 2023.

\bibitem{li2023blip}
J.~Li, D.~Li, S.~Savarese, and S.~Hoi, ``Blip-2: Bootstrapping language-image pre-training with frozen image encoders and large language models,'' in \emph{International conference on machine learning}.\hskip 1em plus 0.5em minus 0.4em\relax PMLR, 2023, pp. 19\,730--19\,742.

\bibitem{openai2023gpt4}
\BIBentryALTinterwordspacing
OpenAI, ``Gpt-4 technical report,'' 2023, accessed: 2023-10-23. [Online]. Available: \url{https://cdn.openai.com/papers/gpt-4.pdf}
\BIBentrySTDinterwordspacing

\bibitem{rombach2022high}
R.~Rombach, A.~Blattmann, D.~Lorenz, P.~Esser, and B.~Ommer, ``High-resolution image synthesis with latent diffusion models,'' in \emph{Proceedings of the IEEE/CVF conference on computer vision and pattern recognition}, 2022, pp. 10\,684--10\,695.

\bibitem{podell2024sdxl}
\BIBentryALTinterwordspacing
D.~Podell, Z.~English, K.~Lacey, A.~Blattmann, T.~Dockhorn, J.~M{\"u}ller, J.~Penna, and R.~Rombach, ``{SDXL}: Improving latent diffusion models for high-resolution image synthesis,'' in \emph{The Twelfth International Conference on Learning Representations}, 2024. [Online]. Available: \url{https://openreview.net/forum?id=di52zR8xgf}
\BIBentrySTDinterwordspacing

\bibitem{kingma2013auto}
D.~P. Kingma, ``Auto-encoding variational bayes,'' \emph{arXiv preprint arXiv:1312.6114}, 2013.

\bibitem{he2016deep}
K.~He, X.~Zhang, S.~Ren, and J.~Sun, ``Deep residual learning for image recognition,'' in \emph{Proceedings of the IEEE conference on computer vision and pattern recognition}, 2016, pp. 770--778.

\bibitem{ronneberger2015u}
O.~Ronneberger, P.~Fischer, and T.~Brox, ``U-net: Convolutional networks for biomedical image segmentation,'' in \emph{Medical image computing and computer-assisted intervention--MICCAI 2015: 18th international conference, Munich, Germany, October 5-9, 2015, proceedings, part III 18}.\hskip 1em plus 0.5em minus 0.4em\relax Springer, 2015, pp. 234--241.

\bibitem{NEURIPS2023_skipgrad}
\BIBentryALTinterwordspacing
Z.~Chen, B.~Li, S.~Wu, K.~Jiang, S.~Ding, and W.~Zhang, ``Content-based unrestricted adversarial attack,'' in \emph{Advances in Neural Information Processing Systems}, A.~Oh, T.~Naumann, A.~Globerson, K.~Saenko, M.~Hardt, and S.~Levine, Eds., vol.~36.\hskip 1em plus 0.5em minus 0.4em\relax Curran Associates, Inc., 2023, pp. 51\,719--51\,733. [Online]. Available: \url{https://proceedings.neurips.cc/paper_files/paper/2023/file/a24cd16bc361afa78e57d31d34f3d936-Paper-Conference.pdf}
\BIBentrySTDinterwordspacing

\bibitem{huang2021mos}
R.~Huang and Y.~Li, ``Mos: Towards scaling out-of-distribution detection for large semantic space,'' in \emph{Proceedings of the IEEE/CVF Conference on Computer Vision and Pattern Recognition}, 2021, pp. 8710--8719.

\bibitem{russakovsky2015imagenet}
O.~Russakovsky, J.~Deng, H.~Su, J.~Krause, S.~Satheesh, S.~Ma, Z.~Huang, A.~Karpathy, A.~Khosla, M.~Bernstein \emph{et~al.}, ``Imagenet large scale visual recognition challenge,'' \emph{International journal of computer vision}, vol. 115, pp. 211--252, 2015.

\bibitem{van2018inaturalist}
G.~Van~Horn, O.~Mac~Aodha, Y.~Song, Y.~Cui, C.~Sun, A.~Shepard, H.~Adam, P.~Perona, and S.~Belongie, ``The inaturalist species classification and detection dataset,'' in \emph{Proceedings of the IEEE conference on computer vision and pattern recognition}, 2018, pp. 8769--8778.

\bibitem{xiao2010sun}
J.~Xiao, J.~Hays, K.~A. Ehinger, A.~Oliva, and A.~Torralba, ``Sun database: Large-scale scene recognition from abbey to zoo,'' in \emph{2010 IEEE computer society conference on computer vision and pattern recognition}.\hskip 1em plus 0.5em minus 0.4em\relax IEEE, 2010, pp. 3485--3492.

\bibitem{zhou2017places}
B.~Zhou, A.~Lapedriza, A.~Khosla, A.~Oliva, and A.~Torralba, ``Places: A 10 million image database for scene recognition,'' \emph{IEEE transactions on pattern analysis and machine intelligence}, vol.~40, no.~6, pp. 1452--1464, 2017.

\bibitem{cimpoi2014describing}
M.~Cimpoi, S.~Maji, I.~Kokkinos, S.~Mohamed, and A.~Vedaldi, ``Describing textures in the wild,'' in \emph{Proceedings of the IEEE conference on computer vision and pattern recognition}, 2014, pp. 3606--3613.

\bibitem{yang2022openood}
J.~Yang, P.~Wang, D.~Zou, Z.~Zhou, K.~Ding, W.~Peng, H.~Wang, G.~Chen, B.~Li, Y.~Sun \emph{et~al.}, ``Openood: Benchmarking generalized out-of-distribution detection,'' \emph{Advances in Neural Information Processing Systems}, vol.~35, pp. 32\,598--32\,611, 2022.

\bibitem{zhang2023openood}
J.~Zhang, J.~Yang, P.~Wang, H.~Wang, Y.~Lin, H.~Zhang, Y.~Sun, X.~Du, K.~Zhou, W.~Zhang \emph{et~al.}, ``Openood v1. 5: Enhanced benchmark for out-of-distribution detection,'' \emph{arXiv preprint arXiv:2306.09301}, 2023.

\bibitem{vaze2021open}
S.~Vaze, K.~Han, A.~Vedaldi, and A.~Zisserman, ``Open-set recognition: A good closed-set classifier is all you need,'' in \emph{International Conference on Learning Representations}, 2021.

\bibitem{bitterwolf2023or}
J.~Bitterwolf, M.~Mueller, and M.~Hein, ``In or out? fixing imagenet out-of-distribution detection evaluation,'' \emph{arXiv preprint arXiv:2306.00826}, 2023.

\bibitem{liu2021swin}
Z.~Liu, Y.~Lin, Y.~Cao, H.~Hu, Y.~Wei, Z.~Zhang, S.~Lin, and B.~Guo, ``Swin transformer: Hierarchical vision transformer using shifted windows,'' in \emph{Proceedings of the IEEE/CVF international conference on computer vision}, 2021, pp. 10\,012--10\,022.

\bibitem{dosovitskiy2021imageworth16x16words}
\BIBentryALTinterwordspacing
A.~Dosovitskiy, L.~Beyer, A.~Kolesnikov, D.~Weissenborn, X.~Zhai, T.~Unterthiner, M.~Dehghani, M.~Minderer, G.~Heigold, S.~Gelly, J.~Uszkoreit, and N.~Houlsby, ``An image is worth 16x16 words: Transformers for image recognition at scale,'' 2021. [Online]. Available: \url{https://arxiv.org/abs/2010.11929}
\BIBentrySTDinterwordspacing

\bibitem{huang2017densely}
G.~Huang, Z.~Liu, L.~Van Der~Maaten, and K.~Q. Weinberger, ``Densely connected convolutional networks,'' in \emph{Proceedings of the IEEE conference on computer vision and pattern recognition}, 2017, pp. 4700--4708.

\bibitem{xie2017aggregated}
S.~Xie, R.~Girshick, P.~Doll{\'a}r, Z.~Tu, and K.~He, ``Aggregated residual transformations for deep neural networks,'' in \emph{Proceedings of the IEEE conference on computer vision and pattern recognition}, 2017, pp. 1492--1500.

\bibitem{paszke2019pytorch}
A.~Paszke, S.~Gross, F.~Massa, A.~Lerer, J.~Bradbury, G.~Chanan, T.~Killeen, Z.~Lin, N.~Gimelshein, L.~Antiga \emph{et~al.}, ``Pytorch: An imperative style, high-performance deep learning library,'' \emph{Advances in neural information processing systems}, vol.~32, 2019.

\bibitem{liu2023gen}
X.~Liu, Y.~Lochman, and C.~Zach, ``Gen: Pushing the limits of softmax-based out-of-distribution detection,'' in \emph{Proceedings of the IEEE/CVF Conference on Computer Vision and Pattern Recognition}, 2023, pp. 23\,946--23\,955.

\bibitem{hendrycks2019augmix}
D.~Hendrycks, N.~Mu, E.~D. Cubuk, B.~Zoph, J.~Gilmer, and B.~Lakshminarayanan, ``Augmix: A simple data processing method to improve robustness and uncertainty,'' \emph{arXiv preprint arXiv:1912.02781}, 2019.

\bibitem{ren2021simple}
J.~Ren, S.~Fort, J.~Liu, A.~G. Roy, S.~Padhy, and B.~Lakshminarayanan, ``A simple fix to mahalanobis distance for improving near-ood detection,'' \emph{arXiv preprint arXiv:2106.09022}, 2021.

\bibitem{djurisic2022extremely}
A.~Djurisic, N.~Bozanic, A.~Ashok, and R.~Liu, ``Extremely simple activation shaping for out-of-distribution detection,'' \emph{arXiv preprint arXiv:2209.09858}, 2022.

\bibitem{hendrycks2021natural}
D.~Hendrycks, K.~Zhao, S.~Basart, J.~Steinhardt, and D.~Song, ``Natural adversarial examples,'' in \emph{Proceedings of the IEEE/CVF conference on computer vision and pattern recognition}, 2021, pp. 15\,262--15\,271.

\bibitem{lee2018simple}
K.~Lee, K.~Lee, H.~Lee, and J.~Shin, ``A simple unified framework for detecting out-of-distribution samples and adversarial attacks,'' \emph{Advances in neural information processing systems}, vol.~31, 2018.

\bibitem{winkens2020contrastive}
J.~Winkens, R.~Bunel, A.~G. Roy, R.~Stanforth, V.~Natarajan, J.~R. Ledsam, P.~MacWilliams, P.~Kohli, A.~Karthikesalingam, S.~Kohl \emph{et~al.}, ``Contrastive training for improved out-of-distribution detection,'' \emph{arXiv preprint arXiv:2007.05566}, 2020.

\bibitem{chen2020boundary}
X.~Chen, X.~Lan, F.~Sun, and N.~Zheng, ``A boundary based out-of-distribution classifier for generalized zero-shot learning,'' in \emph{European conference on computer vision}.\hskip 1em plus 0.5em minus 0.4em\relax Springer, 2020, pp. 572--588.

\bibitem{zaeemzadeh2021out}
A.~Zaeemzadeh, N.~Bisagno, Z.~Sambugaro, N.~Conci, N.~Rahnavard, and M.~Shah, ``Out-of-distribution detection using union of 1-dimensional subspaces,'' in \emph{Proceedings of the IEEE/CVF conference on Computer Vision and Pattern Recognition}, 2021, pp. 9452--9461.

\bibitem{van2020uncertainty}
J.~Van~Amersfoort, L.~Smith, Y.~W. Teh, and Y.~Gal, ``Uncertainty estimation using a single deep deterministic neural network,'' in \emph{International conference on machine learning}.\hskip 1em plus 0.5em minus 0.4em\relax PMLR, 2020, pp. 9690--9700.

\bibitem{huang2020feature}
H.~Huang, Z.~Li, L.~Wang, S.~Chen, B.~Dong, and X.~Zhou, ``Feature space singularity for out-of-distribution detection,'' \emph{arXiv preprint arXiv:2011.14654}, 2020.

\bibitem{fort2021exploring}
S.~Fort, J.~Ren, and B.~Lakshminarayanan, ``Exploring the limits of out-of-distribution detection,'' \emph{Advances in Neural Information Processing Systems}, vol.~34, pp. 7068--7081, 2021.

\bibitem{liu2024visual}
H.~Liu, C.~Li, Q.~Wu, and Y.~J. Lee, ``Visual instruction tuning,'' \emph{Advances in neural information processing systems}, vol.~36, 2024.

\bibitem{zhang2024lapt}
Y.~Zhang, W.~Zhu, C.~He, and L.~Zhang, ``Lapt: Label-driven automated prompt tuning for ood detection with vision-language models,'' in \emph{European Conference on Computer Vision}.\hskip 1em plus 0.5em minus 0.4em\relax Springer, 2024, pp. 271--288.

\bibitem{du2023dream}
X.~Du, Y.~Sun, J.~Zhu, and Y.~Li, ``Dream the impossible: Outlier imagination with diffusion models,'' \emph{Advances in Neural Information Processing Systems}, vol.~36, pp. 60\,878--60\,901, 2023.

\bibitem{chen2024conjugated}
M.~Chen, J.~Gao, and C.~Xu, ``Conjugated semantic pool improves ood detection with pre-trained vision-language models,'' \emph{Advances in Neural Information Processing Systems}, vol.~37, pp. 82\,560--82\,593, 2024.

\bibitem{zhang2024adaneg}
\BIBentryALTinterwordspacing
Y.~Zhang and L.~Zhang, ``Adaneg: Adaptive negative proxy guided {OOD} detection with vision-language models,'' in \emph{The Thirty-eighth Annual Conference on Neural Information Processing Systems}, 2024. [Online]. Available: \url{https://openreview.net/forum?id=vS5NC7jtCI}
\BIBentrySTDinterwordspacing

\bibitem{fang2022out}
Z.~Fang, Y.~Li, J.~Lu, J.~Dong, B.~Han, and F.~Liu, ``Is out-of-distribution detection learnable?'' \emph{Advances in Neural Information Processing Systems}, vol.~35, pp. 37\,199--37\,213, 2022.

\bibitem{zheng2023out}
H.~Zheng, Q.~Wang, Z.~Fang, X.~Xia, F.~Liu, T.~Liu, and B.~Han, ``Out-of-distribution detection learning with unreliable out-of-distribution sources,'' \emph{Advances in neural information processing systems}, vol.~36, pp. 72\,110--72\,123, 2023.

\end{thebibliography}
}

\end{document}